# Investigation of Variances in Belief Networks

Richard E. Neapolitan
Computer Science Dept.
Northeastern Illinois University
Chicago, IL 60625

James R. Kenevan
Computer Science Dept.
Illinois Institute of Technology
Chicago, IL 60616

## Abstract

The belief network is a well-known graphical structure for representing independences in a joint probability distribution. The methods, which perform probabilistic inference in belief networks, often treat the conditional probabilities which are stored in the network as certain values. However, if one takes either a subjectivistic or a limiting frequency approach to probability, one can never be certain of probability values. An algorithm should not only be capable of reporting the probabilities of the alternatives of remaining nodes when other nodes are instantiated; it should also be capable of reporting the uncertainty in these probabilities relative to the uncertainty in the probabilities which are stored in the network. In this paper a method for determining the variances in inferred probabilities is obtained under the assumption that a posterior distribution on the uncertainty variables can be approximated by the prior distribution. It is shown that this assumption is plausible if their is a reasonable amount of confidence in the probabilities which are stored in the network. Furthermore in this paper, a surprising upper bound for the prior variances in the probabilities of the alternatives of all nodes is obtained in the case where the probability distributions of the probabilities of the alternatives are beta distributions. It is shown that the prior variance in the probability at an alternative of a node is bounded above by the largest variance in an element of the conditional probability distribution for that node.

## 1 INTRODUCTION

Much recent research in decision analysis and in expert systems which reason under uncertainty has focused on belief networks. A belief network consists of a DAG $= (V, E)$ in which each $v \in V$ represents a set of mutually exclusive and exhaustive events, along with a joint probability distribution, $P$, on the alternatives of the nodes in $V$. The fundamental assumption in a belief network is that the value assumed by a node is probabilistically independent of the values assumed by all other nodes in the network, except the descendents of the given node, given values of all parents of the node. It can be shown that, given this restriction on $P$, $P$ can be retrieved from the product of the conditional distributions of each node given values of its parents; it can further be shown that, if these conditional distributions are freely specified, the product of those distributions, along with the DAG, constitute a belief network. See Heckerman and Horvitz [1987], Neapolitan [1990], Pearl [1988], and Clemen [1991] for discussions of the importance of belief networks in representing problems in expert systems and decision analysis.

Two important problems in belief networks are probability propagation and abductive inference. Probability propagation is the determination of the values of all other nodes in the network given that certain nodes are instantiated for particular values or that evidence is obtained for the values of certain nodes, while abductive inference is the determination of the most probable, second most probable, third most probable, and so on values of a specified set of nodes called the explanation set given that certain nodes are instantiated or that evidence is obtained. Pearl [1986] and Lauritzen and Spiegelhalter [1988] have obtained efficient algorithms for probability propagation for certain classes of networks, while Cooper [1984], Pearl [1987], and Peng and Reggia [1987] have obtained algorithms which perform abductive inference for certain classes of networks. This paper is concerned only with probabilities obtained using probability propagation. Such probabilities will be called inferred probabilities. The development of efficient general purpose algorithms for probability propagation and abductive inference appears unlikely since Cooper [1988] has shown that both these problems are NP-hard. Recent research has therefore centered on development of approximation, special case, and heuristic methods.

The above methods treat the conditional probabilities which are stored in the network as certain values. For example, Chavez and Cooper's [1990] approximation method computes an error relative to an exact value which would be obtained if exact probability propagation were possible (e.g. using the method of Pearl [1986] or Lauritzen and Spiegelhalter [1988].) However, if one takes either a subjectivistic or a limiting frequency approach to probability, one can never be certain of probability values. Only a pure logical approach claims to know probabilities for certain. For example, if a coin were tossed 1000 times and 527 tosses came up heads, the frequentist would obtain a confidence interval for the probability value, while the subjectivist would obtain a posterior probability interval or a beta posterior distribution.

An exact algorithm for probability propagation should not only be capable of reporting inferred probabilities; it should also be capable of reporting the uncertainty in these probabilities relative to the uncertainty in the conditional probabilities which are specified in the network. An approximation algorithm should incorporate this uncertainty into the possible error which is reported for the approximating values. The uncertainty in the conditional probabilities which are specified in the network can be expressed by probability distributions on the probabilities, and the uncertainty in the inferred probabilities can be determined by computing the variances in these probabilities relative to the joint distribution of these distributions. If a belief network includes decision nodes and a value node, (such a belief network is called an influence diagram, see Clemen [1991]), and the system maximizes expected utility, there are two important reasons for reporting the variances. First, the variances can be used to measure the quality of the system. If a decision is based on a probability of .8 when the 'correct' probability is .4, it may not be a good decision. Second, in an individual case, a large variance may indicate that the best decision would be to gather additional information which would decrease the variance. Howard [1970] discusses variances and the value of information. If the system does not maximize expected utility, the variances in the inferred probabilities inform the decision maker as to the system's uncertainty in its probabilities. This uncertainty should be taken into account before a decision is made.

Results of Zabell [1982] show that in many of the situations involving repeatable experiments the uncertainty in probability values must be represented by Dirichlet distributions. Using a method developed by Spiegelhalter [1988], Neapolitan [1990] showed how to 'discretize' the Dirichlet distributions and represent the uncertainty in the conditional probabilities which are specified in the network in the natural framework of the belief network. For example, in Figure 1 the node $C$ represents the uncertainty in the prior probability of $A$ while $D$ and $E$ represent the uncertainty in the conditional probability of $B$ given $A$. Neapolitan [1990] further showed how to use one of the algorithms for exact probability propagation to compute the variance in an inferred probability relative to the uncertainty in the probabilities which are stored in the network. Neapolitan noted, however, that the number of calculations needed in this computation can grow exponentially with the distance in the graph of a given node from the instantiated node. This is true even in sparsely connected networks for which exact probability propagation is computationally feasible.

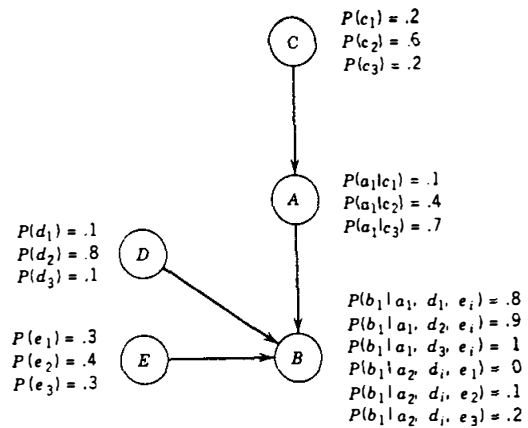

Figure 1: The nodes $C$, $D$, and $E$ represent the uncertainty in probabilities.

Thus there still existed a need for a method for determining the variances in inferred probabilities relative to the uncertainty in the stored probabilities. In Neapolitan and Kenevan [1990] a method is given for determining the prior variances (that is, the variances in the marginal probabilities before any evidence is obtained) of the probabilities of all nodes for the case where certain information is available concerning the probability distributions of the conditional probabilities which are specified in the network. It is also shown how to obtain this information when the distributions are Dirichlet. A method for determining the variances of inferred probabilities appears very difficult even in the case of sparsely connected networks. In Section 3, an exact method is obtained for the case where a posterior distribution on the uncertainty variables can be approximated by the prior distribution. It is shown that this approximation is plausible if there is a reasonable amount of confidence in the probabilities which are stored in the network. Note that, even if there is a reasonable amount of confidence in the stored probabilities, there may still be little confidence in the probabilities of the values of a particular node since the uncertainty in these probabilities is relative to the uncertainty in many of the probabilities which are stored in the network. Thus it is still necessary to compute the variances.

One fear concerning belief networks is that, when a probability value is computed for many of the probabilities which are stored in the network, the variance in that probability value might be hopelessly large since it is relative to the uncertainty in many probabilities. Neapolitan and Kenevan [1990] show that if any probability is very large or very small, then the variance in that probability must be small. This is encouraging since, in medical applications for example, information is often obtained until the probability of some explanation is close to 1. However, what of the case where the probability is not large or small? Applications of algorithm in Neapolitan and Kenevan [1990] indicated that the prior variances did not, as one might expect, become hopelessly large as one went down the network. This indication led to the surprising result which is the main theorem at the end of Section 4 of this paper. Namely that, under certain assumptions, the prior variance in the probability of an alternative of a node is bounded above by the largest variance in an element of the conditional distribution which is specified for that node. Therefore, if we have confidence in the conditional probabilities which are specified in the network, we can have confidence in the prior probabilities of all nodes.

## 2  PRELIMINARY ASSUMPTIONS

It is assumed in what follows that probabilistic assessments in the belief network are made independently. Thus the uncertainties in the assessed probabilities can be represented by a set of mutually independent auxiliary parent nodes. The auxiliary parent of a node, $E$, will be denoted $U_E$. For example, in Figure 2, $U_E$ represents the uncertainty in the $P(E)$, the prior probability of $E$, $U_F$ represents the uncertainty in the $P(F|E)$, the conditional probability of $F$ given $E$, and $U_D$ represents the uncertainty in the $P(D|F,C)$. Each auxiliary node is actually a set of mutually independent nodes, one for each combination of values of the true parents. For example, $U_E$ consists of one node, if $E$ has three alternatives, $U_F$ consists of three nodes, and if $F$ and $C$ each have two alternatives, $U_D$ consists of four nodes. $U$ will be used to denote the set of all the uncertainty nodes. The underlying distribution is then the joint probability distribution on the members of $U$ and will be denoted by $P(U)$. The probabilistic assessments are random variables on this joint probability distribution. Small $p$ will be used to denote these random variables. For example, $p(e_i)$ is the random variable for the prior probability of $e_i$. This random variable will also be denoted by $P(e_i|U)$ when it is convenient to do so. Similarly, $p(f_i|e_j)$ is the random variable for the conditional probability of $f_i$ given that $E$ is equal to $e_j$. This random variable will also be denoted by $P(f_i|e_j,U)$. It is assumed, for example, that $p(e_1)$ is a function only of $U_E$, and that $p(f_1|e_1)$ is a function only of the first member of $U_F$, and $p(f_1|e_2)$ is a function only of the second member of $U_F$. Therefore these random variables are mutually independent. Note however that $p(e_i)$ and $p(e_k)$ are not independent. For example, if $E$ has two alternatives and $p(e_1) = .4$, then $p(e_2)$ must equal .6.

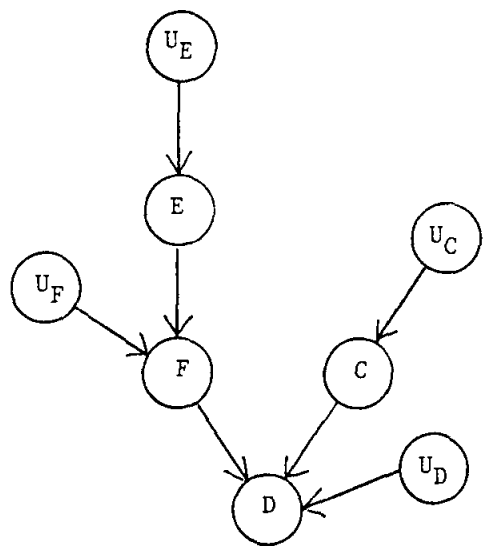

Figure 2: The auxiliary parent nodes represent the uncertainty in probabilities.

The random variable for the probability of a node which is not a root is computed from the assessed random variables. For example,

$$p(f_1) = \sum_i p(f_1|e_i)p(e_i)$$

$$p(d_1) = \sum_{i,k} p(d_1|f_i,c_k)p(f_i,c_k)$$
$$= \sum_{i,k} p(d_1|f_i,c_k)p(f_i)p(c_k)$$

since $p(f_i)$ and $p(c_k)$ are independent random variables due to the network being singly connected.

If $p_i$ is a random variable for a probability value which is stored in the network (e.g., $p_i$ may be $p(e_i)$ or $p(f_i,e_j)$), it is assumed in this paper that the following information is available for $p_i$, where $E$ stands for the expected value:

$$E(p_i) \qquad E(p_i^2) \qquad E(p_ip_j) \qquad (1)$$

Note that $p_i$ and $p_j$ are random variables for the $i^{th}$ and $j^{th}$ alternatives of the same node. Neapolitan and Kenevan [1990] obtained this information in the case where the distributions are Dirichlet.

## 3 DETERMINING THE VARIANCES IN INFERRED PROBABILITIES

In this section a method is obtained for computing the posterior variances (i.e., the variances in the inferred probabilities given that certain nodes, $W$, are instantiated) under the assumption that a posterior distribution on the uncertainty variables, $P(U|V)$, can be approximated by the prior distribution, $P(U)$. Neapolitan [1991] shows that, when nodes are instantiated, the posterior distribution on the uncertainty variables can differ from the prior distribution by no more than a distribution based on the information in one additional trial. Therefore if there is a reasonable amount of confidence in the prior distribution, a posterior distribution can be approximated by the prior distribution. Essentially, we are assuming that we learn nothing about the probabilities which are specified in the network from the current case.

In this section, it is assumed that a variable has exactly two sons for the sake of clarity. The case of an arbitrary number of sons is a straightforward generalization. Furthermore, if $W$ is the set of instantiated variables, $E_W$ will be used to represent an expected value relative to the posterior distribution on the uncertainty variables, while $E$ will continue to represent an expected value relative to the prior distribution.

### 3.1 THE CASE OF TREES

If $W$ is a set of instantiated nodes, we are interested in the variance of the probability of an uninstantiated node, $F$, relative to the posterior distribution on the uncertainty variables. This variance is given by

$$V_W(p(f_i|W)) = E_W(p(f_i|W)^2) - (E_W(p(f_i|W)))^2.$$

Thus our goal is to compute these latter two quantities. Since

$$E_W(p(f_i|W)) = \int_U P(f_i|W,U)dP(U|W) = P(f_i|W),$$

we could compute $E_W(p(f_i|W))$ using one of the standard methods for computing probabilities. However, the method described here is a generalization of Pearl's [1986] method for computing probabilities, and the determination of $P(f_i|W)$ is a by-product of the determination of $E(p(f_i|W)^2)$. The propagation scheme is based on the following theorem.

**Theorem 1** *Let $F$ be an uninstantiated variable, $W$ be the set of instantiated variables, $X_F$ be the set of instantiated variables in the tree rooted at $F$'s left son, $V_F$ be the set of instantiated variables in the tree rooted at $F$'s right son, and let $Y_F = W - X_F - V_F$. Then, with the assumption that a posterior distribution on the uncertainty variables can be approximated by the prior distribution, it is the case that*

$$E_W(p(f_i|W)^2) =$$
$$\frac{E(p(X_F|f_i)^2)E(p(V_F|f_i)^2)E(p(f_i|Y_F)^2)}{\left(\sum_j P(X_F|f_j)P(V_F|f_j)P(f_j|Y_F)\right)^2}$$

*and*

$$E_W(p(f_i|W)) = P(f_i|W)$$
$$= \frac{P(X_F|f_i)P(V_F|f_i)P(f_i|Y_F)}{\sum_j P(X_F|f_j)P(V_F|f_j)P(f_j|Y_F)}.$$

**Proof** Let $U$ be the set of uncertainty variables, $U_X$ be the subset of those variables which are connected to $F$ through $F$'s left son, $U_V$ be the subset which is connected to $F$ through $F$'s right son, and $U_Y$ be the remainder of the uncertainty variables. Then

$$E_W(p(f_i|W)^2) = \int_U P(f_i|W,U)^2 dP(U|W)$$
$$= \int_U \frac{P(W|f_i,U)^2 P(f_i|U)^2}{P(W|U)^2} dP(U|W)$$
$$= \int_U \frac{P(W|f_i,U)^2 P(f_i|U)^2 dP(U)^2}{dP(U|W)^2 P(W)^2} dP(U|W)$$
$$= \int_U \frac{P(W|f_i,U)^2 P(f_i|U)^2 dP(U)}{dP(U|W)P(W)^2} dP(U)$$
$$= \int_U \frac{P(W|f_i,U)^2 P(f_i|U)^2 dP(U)}{dP(U)P(W)^2} dP(U)$$

The second to the last equality is due to the assumption that a posterior distribution on uncertainty variables can be approximated by the prior distribution. Thus we have that

$$E_W(p(f_i|W)^2 =$$
$$\frac{P(Y_F)^2}{P(W)^2} E(p(X_F|f_i)^2 E(p(V_F|f_i)^2)E(p(f_i|Y_F)^2).$$

Finally,

$$\frac{P(Y_F)}{P(W)} = \frac{P(Y_F)}{P(X_F,V_F,Y_F)} = \frac{P(Y_F)}{P(X_F,V_F|Y_F)P(Y_F)}.$$

Again due to d-separation

$$P(X_F,V_F|Y_F) = \sum_j P(X_F,V_F|Y_F,f_j)P(f_j|Y_F)$$
$$= \sum_j P(X_F|f_j)P(V_F|f_j)P(f_j|Y_F),$$

which proves the first part of the theorem. The second part is proved in the same fashion and does not require the assumption that a posterior distribution on the uncertainty variables can be approximated by the prior distribution. □

Due to Theorem 1, the information which $F$ needs from its parent is

$$P(f_i|Y_F) \quad \text{and} \quad E(p(f_i|Y_F)^2),$$

while the information which $F$ needs from its sons is

$$P(X_F|f_i), \quad E(p(X_F|f_i)^2),$$
$$P(V_F|f_i), \quad E(p(V_F|f_i)^2).$$

First we will show how to obtain the information which $F$ needs from its sons. Suppose that the following information is available for each son, $G$, of $F$:

$$P(X_G|g_i), \quad E(p(X_G|g_i)^2),$$
$$E(p(X_G|g_i)(p(X_G|g_j)),$$
$$P(V_G|g_i), \quad E(p(V_G|g_i)^2),$$
$$E(p(V_G|g_i)p(V_G|g_j)) \quad (2)$$

Notice that $p(X_G|g_i)$ and $p(X_G|g_j)$ are not in general independent since they both depend on uncertainty variables in the tree rooted at $G$. Suppose that $G$ is the left son of $F$. Then the tree rooted at $G$ contains $X_F$. We shall first consider the case where $G$ is not instantiated. In that case we have, due to d-separation, that

$$E(p(X_F|g_i)^2) = E(p(X_G, V_G|g_i)^2)$$
$$= E(p(X_G|g_i)^2)E(p(V_G|g_i)^2)$$

and similarly that

$$P(X_F|g_i) = P(X_G|g_i)P(V_G|g_i)$$
$$E(p(X_F|g_i)p(X_F|g_j)) =$$
$$E(p(X_G|g_i)p(X_G|g_j))E(p(V_G|g_i)p(V_G|g_j)).$$

Thus if the information listed in (2) is available for $G$ we can compute $E(p(X_F|g_i)^2)$, $P(X_F|g_i)$, and $E(p(X_F|g_i)p(X_F|g_j))$. We will now show that the first half of the information listed in (2) for $F$ can be computed from this latter information. First we have that

$$P(X_F|f_i) = \sum_j P(X_F|g_j)P(g_j|f_i).$$

Since $P(g_j|f_i) = E(p(g_j|f_i))$, and this value is part of the information listed in (1) for $p(g_j|f_i)$, we see that we can compute $P(X_F|f_i)$ from the information listed in (2) for $G$ and the information listed in (1) for distributions which are specified in the network. Next we have that

$$E(p(X_F|f_i)p(X_F|f_j)) =$$
$$E\left(\left(\sum_k p(X_F|g_k)p(g_k|f_i)\right)\left(\sum_k p(X_F|g_k)p(g_k|f_j)\right)\right)$$

This latter expression is the sum of the following kind of terms:

$$E(p(X_F|g_k)^2)E(p(g_k|f_i))E(p(g_k|f_j))$$
and
$$E(p(X_F|g_k)p(X_F|g_r))E(p(g_k|f_i))E(p(g_r|f_j)).$$

The values in these terms are either part of the information listed in (1) for distributions which are specified in the network or, as previously shown, can be computed from the information listed in (2) for $G$. Finally

$$E\left(p(X_F|f_i)^2\right) =$$
$$E\left(\left(\sum_k p(X_F|g_k)p(g_k|f_i)\right)\left(\sum_k p(X_F|g_k)p(g_k|f_i)\right)\right)$$

This latter expression contains the following kinds of terms:

$$E\left(p(X_F|g_k)^2\right)E\left(p(g_k|f_i)^2\right)$$
$$E(p(X_F|g_k)p(X_F|g_j))E(p(g_k|f_i)p(g_j|f_i)).$$

Therefore $E(p(X_F|f_i)^2)$ can also be computed from the information listed in (2) for $G$ and the information listed in (1) for distributions which are specified in the network.

Next suppose that $G$ is instantiated for $g_1$. In this case, due to d-separation, we have that

$$P(X_F|f_i) = P(X_G, V_G, g_1|f_i)$$
$$= P(X_G|V_G, g_1, f_i)P(V_G|g_1, f_i)P(g_1, f_i)$$
$$= P(X_G|g_1)P(V_G|g_1)P(g_1|f_i).$$

Similarly

$$E\left(p(X_F|f_i)^2\right) =$$
$$E\left(p(X_G|g_1)^2\right)E\left(p(V_G|g_1)^2\right)E\left(p(g_1|f_i)^2\right)$$
$$E(p(X_F|f_i)p(X_F|f_j)) =$$
$$E\left(p(X_G|g_1)^2\right)E\left(p(V_G|g_1)^2\right)$$
$$E(p(g_1|f_i))E(p(g_1|f_j)).$$

Thus again the information listed in (2) for $F$ can be computed from the information listed in (2) for the sons of $F$ along with the information listed in (1) for distributions which are specified in the network.

Since $G$ d-separates $F$ from the uncertainty variables connected to $F$ through $G$, it does not seem correct, for example, that $E(p(f_i|W)^2)$ depends on $E(p(X_G|g_1)^2)$ and $E(p(V_G|g_1)^2)$ when $G$ is instantiated for $g_1$. This apparent dependence was caused by the assumption that a posterior distribution on the uncertainty variables can be approximated by the prior distribution. As shown in Neapolitan [1991] this dependence does not really exist, and when $F$'s son, $G$, is instantiated for $g_1$, $X_F$ can be set equal to $\{g_1\}$ in Theorem 3.1 and

$$P(X_F|f_i) = P(g_1|f_i)$$
$$E\left(p(X_F|f_i)^2\right) = E\left(p(g_1|f_i)^2\right)$$
$$E(p(X_F|f_i)p(X_F|f_j)) = E(p(g_1|f_i))E(p(g_1|f_j)).$$

These latter values are simply the information listed in (1) for distributions which are specified in the network.

Next we show how to obtain the information which a node, $F$, needs from its parent, $E$. We will show that the following information can be obtained from information listed in (1) for distributions which are stored in the network if $E$ is instantiated or, if $E$ is not instantiated, from the corresponding information for $E$ along with the information listed in (2) for $E$ which comes from $E$'s son (since we are assuming that a posterior distribution on the uncertainty variables can be approximated by the prior distribution, notice that this information is the information listed in (1) conditional on $Y_F$):

$$P(f_i|Y_F), \quad E\left(p(f_i|Y_F)^2\right),$$
$$E(p(f_i|Y_F)p(f_j|Y_F)). \qquad (3)$$

Since $E$ d-separates $F$ from all of $F$'s ancestors and $E$'s descendents through $E$'s other children, if $E$ is instantiated for $e_1$ this information is simply equal to

$$P(f_i|e_i) \quad E\left(p(f_i|e_1)^2\right) \quad E(p(f_i|e_1)p(f_j|e_1)),$$

which is the information listed in (1) for distributions which are specified in the network. Next assume that $E$ is not instantiated and the information listed in (2) which comes from $E$'s other son and the information listed in (3) is available for $E$. Due to d-separation

$$P(f_i|Y_F) = \sum_j P(f_i|e_j)P(e_j|Y_F).$$

Assuming that $F$ is the right son of $E$ and that therefore the tree rooted at $F$ contains $V_E$, in the same way that Theorem 1 is proved it is possible to show that

$$P(e_j|Y_F) = \frac{P(X_E|e_j)P(e_j|Y_E)}{\sum_m P(X_E|e_m)P(e_m|Y_E)},$$

and therefore $P(f_i|Y_F)$ can be computed from the information listed in (2) which comes from $E$'s other son and the information listed in (3) for $E$ along with the information listed in (1) for distributions which are stored in the network. Next we have that

$$E\left(p(f_i|Y_F)^2\right) = E\left(\left(\sum_j p(f_i|e_j)p(e_j|Y_F)\right)^2\right).$$

This latter expression is the sum of the following kinds of terms:

$$E\left(p(f_i|e_j)^2\right) E\left(p(e_j|Y_F)^2\right)$$

and

$$E(p(f_i|e_j))E(p(f_i|e_k))E(p(e_j|Y_F)p(e_k|Y_F)).$$

In the same way that Theorem 1 is proved it is possible to show that

$$E\left(p(e_j|Y_F)^2\right) =$$

$$\frac{E\left(p(X_E|e_j)^2\right) E\left(p(e_j|Y_E)^2\right)}{\left(\sum_m P(X_E|e_m)P(e_m|Y_E)\right)^2}$$

$$E(p(e_j|Y_F)p(e_k|Y_F)) =$$
$$\frac{E(p(X_E|e_j)p(X_E|e_k))E(p(e_j|Y_E)p(e_k|Y_E))}{\left(\sum_m P(X_E|e_m)P(e_m|Y_E)\right)^2}.$$

Thus $E(p(f_i|Y_F)^2)$ can be computed from the information listed in (2) for $E$ which comes from $E$'s other son and the information list in (3) for $E$ along with the information listed in (1) for distributions which are specified in the network. Finally

$$E(p(f_i|Y_F)(p(f_i|Y_F)) =$$
$$E\left(\left(\sum_m p(f_i|e_m)p(e_m|Y_F)\right)\right.$$
$$\left.\left(\sum_m p(f_j|e_m)p(e_m|Y_F)\right)\right).$$

This latter expression includes the following types of terms:

$$E(p(f_i|e_m)p(f_j|e_m))E(p(e_m|Y_F)^2)$$
and
$$E(p(f_i|e_m))E(p(f_j|e_r))E(p(e_m|Y_F)p(e_r|Y_F)).$$

We have just shown that the information in these terms can be computed from the information listed in (2) which comes from $E$'s other son and the information listed in (3) for $E$ along with the information listed in (1) for distributions which are specified in the network.

Using the theory developed above we can determine variances as follows. First initialize for every node, $F$, the values of

$$P(X_F|f_i), \quad E(p(X_F|f_i)^2),$$
$$E(p(X_F|f_i)p(X_F|f_j)),$$
$$P(V_F|f_i), \quad E(p(V_F|f_i)^2),$$
$$E(p(V_F|f_i)p(V_F|f_j))$$

all to 1. Then since for the root, $A$,

$$P(a_i|\Phi_A) = P(a_i),$$
$$E(p(a_i|\Phi_A)^2) = E(p(a_i)^2),$$
$$E(p(a_i|\Phi_A)p(a_j|\Phi_A)) = E(p(a_i)p(a_j)),$$

and these values are the information listed in (1) for distributions which are specified in the network, the prior variances can be computed by initiating a propagation flow from the root. At each node the variances are computed using the formulas in Theorem 1. This result agrees with the variances obtained using the exact method described in Neapolitan and Kenevan [1990]. When a node, $G$, is instantiated, it initiates new propagation down by using the method described above for obtaining the information listed in (3) for each son of $G$ from the son's parent in the case were the parent is instantiated. $G$'s uninstantiated children then continue the propagation flow using the method for the case where the parent is uninstantiated. (Note that instantiated nodes are dead ends for downward propagation.) $G$ also initiates new propagation up by using the method described above to send its parent, $F$, the portion of the information listed in (2) for $F$ which comes from $G$. As shown above, this information is stored in the network. If $F$ is uninstantiated, the information listed in (2) for $F$'s parent, which comes from $F$, is then computed using the method described above for uninstantiated nodes. Instantiated nodes are also dead ends for upward propagation. When an uninstantiated node receives new information from below it not only must send new information up but also must send new information down to each of its other sons using the method described above for obtaining the information in (3) in the case where the parent is not instantiated.

### 3.2 THE CASE OF SINGLY CONNECTED AND ARBITRARY BELIEF NETWORKS

The method described above can be extended to the case of singly connected networks. This extension appears in Neapolitan [1991]. The case of an arbitrary network can then be handled by using Pearl's [1988] method of clustering as discussed in Neapolitan and Kenevan [1990].

## 4 OBTAINING AN UPPER BOUND FOR THE PRIOR VARIANCES

As noted in the introduction, Zabell [1982] has shown that in many of the cases which are relevant to expert systems, the probability distribution of a probability must be Dirichlet. In this section it is assumed that there are two alternatives for each node and the distributions stored in the network are all Dirichlet. In the case of two alternatives, the Dirichlet distribution is called the beta distribution. It is given by

$$\mu(p) = \frac{(a+b+1)!}{a!b!} p^a (1-p)^b,$$

where $a$ and $b$ are nonnegative parameters.

It is straightforward to show that, in the case of the beta distribution,

$$E(p) = \frac{a+1}{a+b+2}$$
$$E(p^2) = \frac{a+2}{a+b+3} E(p)$$
$$E(p(1-p)) = \frac{b+1}{a+b+3} E(p) \quad (4)$$

where $E$ stands for expected value. Some of these results are needed to obtain the proofs in this section.

The main theorem at the end of this section obtains the upper bounds for the prior variances of the probabilities of all nodes in the network in the case where the network is a tree and the distributions stored in the network are beta. First we must obtain a number of preliminary results.

**Lemma 1** *If there are exactly two alternatives, then $V(p_1) = V(p_2)$, however if there are three or more alternatives, the variances are not in general equal.*

**Proof** Since

$$V(p_1) = \int_U p_1^2 dP(U) - \left( \int_U p_1 dP(U) \right)^2$$

and

$$V(p_2) = \int_U (1-p_1)^2 dP(U) - \left( \int_U (1-p_1) dP(U) \right)^2,$$

the first part of the lemma is proved with simple algebraic manipulations. A counter example using a Dirichlet distribution proves the second part. □

**Lemma 2** *Suppose there are exactly two alternatives. Let $E = E(p_1)$, $S = E(p_1^2)$, and $T = E(p_2^2)$. Then $T = 1 - 2E + S$.*

**Proof** Due to Lemma 1, $V(p_1) = V(P_2)$. Set $V$ be that variance. Since $S = E^2 + V$ and $T = (1-E)^2 + V$, the lemma is proved with straightforward algebraic manipulations. *Box*

**Lemma 3** *Suppose there are exactly two alternatives. Let $P = E(p_1 p_2)$, and $E$ and $S$ be as in Lemma 2. Then $P = E - S$.*

**Proof**

$$\begin{aligned} P &= \int_U p_1(1-p_1) dP(U) \\ &= \int_U p_1 dP(U) - \int_U p_1^2 dP(U) \\ &= E - S. \quad \square \end{aligned}$$

**Lemma 4** *Suppose there are exactly two alternatives for both node $A$ and node $B$ and that $A$ is the only parent of $B$. Let*

$$E = E(a_1), \qquad E_1 = E(p(b_1|a_1)),$$
$$E_2 = E(p(b_1|a_2)),$$
$$V = V(a_1), \qquad V_1 = V(p(b_1|a_1)),$$
$$V_2 = V(p(b_1|a_2))$$

*Then*

$$V(p(b_1)) =$$
$$V(V_1 + V_2 + (E_1 - E_2)^2) + V_2(1-E)^2 + V_1 E^2.$$

**Proof** Let $S = E(p(a_1)^2)$, $T = E(p(a_2)^2)$, $P = E(p(a_1)p(a_2))$, $S_1 = E(p(b_1|a_1)^2)$, and $S_2 = E(p(b_1|a_2)^2)$. We then have

$$E\left(p(b_1)^2\right) = E\left([p(b_1|a_1)p(a_1) + p(b_1|a_2)p(a_2)]^2\right)$$

which is easily seen to equal

$$E\left(p(b_1|a_1)^2\right) E\left(p(a_1)^2\right)$$
$$+ 2E(p(b_1|a_1))E(p(b_1|a_2))E(p(a_1)p(a_2))$$
$$+ E\left(p(b_1|a_2)^2\right) E\left(p(a_2)^2\right),$$

and thus

$$E\left(p(b_1)^2\right) = S_1 S + 2E_1 E_2 P + S_2 T. \quad (5)$$

Similarly it can be shown that

$$(E(p(b_1)))^2 = E_1^2 E^2 + 2E_1 E E_2 - 2E_1 E_2 E^2$$
$$+ E_2^2 - 2E_2^2 E + E_2^2 E^2. \quad (6)$$

Since

$$V(p(b_1)) = E\left(p(b_1)^2\right) - (E(p(b_1)))^2,$$

equations (5) and (6), applications of Lemmas 2 and 3, and some algebraic manipulations yield

$$V(p(b_1)) = V_2 - 2E_1 E_2 V - 2E V_2 + S_1 S$$
$$- E_1^2 E^2 + S_2 S - E_2^2 E^2. \quad (7)$$

Now

$$S_1 S - E_1^2 E^2 = S_1 S - S_1 E^2 + S_1 E^2 - E_1^2 E^2$$
$$= S_1 V + E^2 V_1,$$

and an identical result holds for $S_2 S - E_2^2 E^2$. After inserting these results in equation (7) and performing some more algebraic manipulations we have that

$$V(p(b_1)) = V(S_1 + S_2 - 2E_1 E_2)$$
$$+ V_2(1 - E)^2 + V_1 E^2. \quad (8)$$

Replacing $S_1$ and $S_2$ in (8) by $V_1 + E_1^2$ and $V_2 + E_2^2$ respectively proves the lemma. □

**Lemma 5** *Let $E$, $S$, and $V$ be as in Lemma 2. Then $S \leq (E + E^2)/2$ implies that $V \leq E - S$.*

**Proof** $V \leq E - S$ means $S - E^2 \leq E - S$, which is true if $2S \leq E + E^2$, which is the condition in the statement of the lemma. □

**Lemma 6** *Let $E$ and $S$ be as in Lemma 2. If the distribution is beta, then*

$$S \leq (E + 2E^2)/3.$$

**Proof** Let $P$ be as in Lemma 3, and $a$ and $b$ be the parameters for the beta distribution. Then, due to equalities (4),

$$\begin{aligned} P &= \left(\tfrac{b+1}{a+b+3}\right) E \\ &= \left(\tfrac{a+b+2}{a+b+3}\right)\left(\tfrac{b+1}{a+b+2}\right) E \\ &= \left(\tfrac{a+b+2}{a+b+3}\right)(1 - E)E \\ &\geq 2\left(E - E^2\right)/3. \end{aligned}$$

Thus due to Lemma 3 we have that $E - S \geq 2(E - E^2)/3$, and the lemma follows from some algebraic manipulations. □

**Lemma 7** *If the distribution is beta, then $E \leq (1 + \sqrt{1 - 12V})/2$.*

**Proof** In the same way that Lemma 5 was proved, Lemma 6 implies that

$$V \leq \left(E - E^2\right)/3,$$

which implies

$$E^2 - E + 3V \leq 0.$$

The expression $E^2 - E + 3V$ equals 0 at the point $E = (1 + \sqrt{1 - 12V})/2$ and has positive derivative with respect to $E$ at his point and to the right of it. Therefore if $E^2 - E + 3V \leq 0$, $E$ must be $\leq (1 + \sqrt{1 - 12V})/2$. □

**Theorem 2** *Assume the conditions and notations in Lemma 4. Further assume that $S \leq (E + E^2)/2$. Then*

$$V(p(b_1)) \leq Maximum(V_1, V_2).$$

**Proof** Suppose $V_1 \leq V_2$. Without loss of generality we can assume $E_1 \leq E_2$. For, if this were not the case, we would have

$$E(p(b_2|a_1)) = 1 - E_1 < 1 - E_2 = E(p(b_2|a_2)).$$

However, due to Lemma 1 and the assumption that $V_1 \leq V_2$,

$$\begin{aligned} V(p(b_2|a_1)) &= V(p(b_1|a_1)) = V_1 \\ &\leq V_2 = V(p(b_1|a_2)) = V(p(b_2|a_2)), \end{aligned}$$

and we could proceed in the proof using $b_2$ instead of $b_1$. Assuming now that $E_1 \leq E_2$, due to Lemma 4 we have that

$$V(p(b_1)) =$$
$$V\left(V_1 + V_2 + (E_2 - E_1)^2\right) + V_2(1 - E)^2 + V_1 E^2$$
$$\leq V\left(2V_2 + E_2^2\right) + V_2(1 - E)^2 + V_2 E^2$$
$$\leq V\left(2V_2 + (1 + \sqrt{1 - 12V_2})/2\right)$$
$$\quad + V_2(1 - E)^2 + V_2 E^2.$$

The last inequality is due to Lemma 7 and the fact that $E_2^2 \leq E_2$. After some algebraic manipulations we have

$$V(p(b_1))$$
$$\leq V(1 + \sqrt{1 - 12V_2})/2$$
$$\quad + V_2(1 + 2V - 2E + 2E^2)$$
$$\leq V(1 + \sqrt{1 - 12V_2})/2$$
$$\quad + V(1 + 2S - 2E^2 - 2E + 2E^2)$$
$$\leq V(1 + \sqrt{1 - 12V_2})/2 + V_2(1 - 2(E - S)).$$

Due to Lemma 5 and the assumptions of this theorem we then have
$$V(p(b_1)) \leq V(1 + \sqrt{1 - 12V_2})/2 + V_2(1 - 2V).$$
Thus $V(p(b_1)) \leq V_2$ if
$$V(1 + \sqrt{1 - 12V_2})/2 + V_2(1 - 2V) \leq V_2.$$
If $V = 0$, the proof is trivial. Thus this latter inequality is true if $1 + \sqrt{1 - 12V_2} \leq 4V_2$, which is true if $1 - 12V_2 \leq 16V_2^2 - 8V_2 + 1$. Since $16V_2^2 + 4V_2 \geq 0$ the theorem is proved. □

**Theorem 3** *Assume the conditions and notation in Lemma 4. Furthermore let $S' = E(p(b_1)^2)$ and $E' = E(p(b_1))$. Then $S \leq (E + E^2)/2$ implies that $S' \leq (E' + E'^2)/2$.*

**Proof** Equation (5) from the proof of Lemma 4 states that
$$S' = S_1 S + 2E_1 E_2 P + S_2 T.$$
It is straightforward that
$$(E' + E'^2) = [E_1 E + E_2(1 - E) + (E_1 E + E_2(1 - E))^2]/2.$$
Thus we need show that
$$S_1 S + 2E_1 E_2 P + S_2 T \leq [E_1 E + E_2(1 - E) + (E_1 E + E_2(1 - E))^2]/2.$$
Algebraic manipulations yield that this is equivalent to showing that
$$(S_1 S + E_1 E_2 P) + (S_2 T + E_1 E_2 P) \leq [E_1 E + E_1^2 E^2 + E_1 E E_2(1 - E)]/2$$
$$+ [E_2(1 - E) + E_2^2(1 - E)^2 + E_1 E E_2(1 - E)]/2.$$
We will accomplish this by showing that
$$S_1 S + E_1 E_2 P \leq$$
$$[E_1 E + E_1^2 E^2 + E_1 E E_2(1 - E)]/2. \quad (9)$$
By symmetry we will then also have shown that
$$S_2 T + E_1 E_2 P \leq$$
$$[E_2(1 - E) + E_2^2(1 - E)^2 + E_1 E E_2(1 - E)]/2,$$
and the theorem will be proved. To that end, due to Lemmas 3 and 6,
$$S_1 S + E_1 E_2 P = S_1 S + E_1 E_2 (E - S)$$
$$\leq (E_1 + 2E_1^2)S/3 + E_1 E_2 E - E_1 E_2 S.$$
Thus inequality (9) will hold if
$$S(E_1 + 2E_1^2 - 3E_1 E_2)/3 + E_1 E_2 E$$
$$\leq [E_1 E + E_1^2 E^2 + E_1 E E_2(1 - E)]/2.$$
Eliminating the trivial case when $E_1 = 0$, algebraic manipulations yield that this last inequality is equivalent to
$$2S(1 + 2E_1 - 3E_2) \leq 3E(1 + E_1 E - EE_2 - E_2). \quad (10)$$

There are two cases:

**Case 1:** $1 + 2E_1 - 3E_2 \geq 0$. In this casse due to the assumption that $S \leq (E + E^2)/2$, inequality (10) is true if
$$(E + E^2)(1 + 2E_1 - 3E_2) \leq$$
$$3E(1 + E_1 E - EE_2 - E_2),$$
which, after performing algebraic manipulations and eliminating the trivial case when $E = 0$, is equivalent to $E(1 - E_1) \leq 2(1 - E_1)$ which proves Case 1.

**Case 2:** $1 + 2E_1 - 3E_2 < 0$. Inequality (10) is equivalent to
$$3E(EE_2 + E_2 - 1 - E_1 E) \leq 2S(3E_2 - 2E_1 - 1),$$
where the right side of this inequality is now positive due to the assumption in this case. Since $V \geq 0$, we have $S \geq E^2$ and therefore the last inequality is true if
$$3E(EE_2 + E_2 - 1 - E_1 E) \leq$$
$$2E^2(3E_2 - 2E_1 - 1).$$
After the trivial case when $E = 0$ is eliminated, algebraic manipulations yield that this inequality is equivalent to
$$2E(1 - E_2) + E(E_1 - E_2) \leq 3(1 - E_2).$$
Clearly $2E(1 - E_2) \leq 2(1 - E_2)$. Therefore we need only show that $E(E_1 - E_2) \leq 1 - E_2$. Due to the assumption in this case, $E_1 < (3E_2 - 1)/2$. Thus we need only show that
$$E((3E_2 - 1)/2 - E_2) \leq 1 - E_2.$$
Algebraic manipulations yields that this is equivalent to $0 \leq (1 - E_2)(2 + E)$, which proves this case. □

**Theorem 4 (Main Theorem)** *If the belief network is a tree, and all distributions stored in the network are beta, then for any node, $B$,*
$$V(p(b_1)) \leq Maximum(V_1, V_2), \quad (11)$$
*where $V_1 = V(p(b_1|a_1))$, $V_2 = V(p(b_1|a_2))$, and $A$ is $B$'s parent.*

**Proof** Due to Lemma 6 and the fact that the distribution for the probability of the root is assumed to be beta, it is easy to show that for the root
$$S \leq (E + E^2)/2. \quad (12)$$
where $S$ and $E$ stand again for the expected value of a probability and the expected value of a probability squared respectively. Thus, due to Theorem 3 and an inductive argument, inequality (12) holds for all nodes in the network. Therefore, due to Theorem 2, inequality (11) holds for all nodes in the network. □

## 5  FUTURE RESEARCH

There exists a need for a great deal of additional research in this area. First, it should be determined if the results in Section 4 hold when there are more than two alternatives and when the network is not tree. Second, of greater interest than the variance in prior probabilities are the variances in inferred probabilities. If a node is instantiated in a tree, it is easy to see that the results above hold for the descendents of that node. However, in many cases we are more interested in the ancestors of the node. In medicine, for example, instantiated nodes, which represent findings, are often near leaves while the nodes of interest, which represent diseases, are often near roots. It remains to be determined whether the variances become large as we propagate up the network from an instantiated node.